\documentclass[12pt]{article}
	\usepackage[margin=1in]{geometry}
	\usepackage[slantedGreek]{mathpazo}
	\usepackage[pdftex]{graphicx}
	\usepackage{amsfonts}
	\usepackage{amssymb}
	\usepackage{amsthm}
	\usepackage{graphicx,float}
	\usepackage{amsmath}%
    \usepackage{subcaption} 
    \usepackage[ruled,vlined]{algorithm2e} 
    \usepackage{natbib} 
	\usepackage{hyperref}
    \usepackage{accents}
\usepackage{color,soul,marginnote}

\newtheorem{remark}{Remark}[section]

\title{Horseshoe Mixtures-of-Experts (HS-MoE)}
\author{	
    \makebox[.4\linewidth]{Nick Polson}\\\textit{Booth School of Business}\\\textit{University of Chicago}\\\and 
    \makebox[.4\linewidth]{Vadim Sokolov\footnote{Nick Polson is at Chicago Booth: ngp@chicagobooth.edu. Vadim Sokolov is Associate Professor at Volgenau School of Engineering, George Mason University, USA: vsokolov@gmu.edu.}}\\\textit{Department of Systems Engineering}\\\textit{and Operations Research}\\\textit{George Mason University}
}
\date{First Draft: December 9, 2025\\This Draft: \today}

\graphicspath{{./fig/}}

\begin{document}
\maketitle

\begin{abstract}
\noindent Horseshoe mixtures-of-experts (HS-MoE) models provide a Bayesian framework for sparse expert selection in mixture-of-experts architectures. We combine the horseshoe prior's adaptive global-local shrinkage with input-dependent gating, yielding data-adaptive sparsity in expert usage. Our primary methodological contribution is a particle learning algorithm for sequential inference, in which the filter is propagated forward in time while tracking only sufficient statistics. We also discuss how HS-MoE relates to modern mixture-of-experts layers in large language models, which are deployed under extreme sparsity constraints (e.g., activating a small number of experts per token out of a large pool).
\end{abstract}


\newpage
\section{Introduction}\label{sec:intro}

Mixture-of-Experts (MoE) models provide a powerful framework for combining multiple specialized models through input-dependent gating \citep{jacobs1991adaptive, jordan1994hierarchical}. The foundational model represents the predictive distribution as:
\begin{equation}\label{eq:moe-basic}
p(Y \mid X, \phi) = \sum_{k=1}^{K} g_k(X; \phi) \, f_k(Y \mid X; \theta_k),
\end{equation}
where $g_k(X; \phi)$ is the gating function assigning input-dependent weights to expert $k$, and $\theta_k$ parameterizes the $k$-th expert. Unlike standard finite mixtures with fixed weights, the input-dependent gating enables adaptive partitioning of the input space \citep{titterington1985statistical}.

\citet{jiang1999hierarchical} established theoretical foundations for hierarchical MoE, including conditions for identifiability and convergence rates. Bayesian approaches to MoE inference were developed by \citet{peng1996bayesian}, who introduced MCMC methods for posterior computation in MoE and hierarchical MoE models.

Recent work has revived MoE in large language models. Sparsely gated MoE layers \citep{shazeer2017outrageously} scale model capacity while keeping per-token computation tractable. The Switch Transformer \citep{fedus2022switch} simplified routing to a single expert per token. Modern architectures such as Mixtral \citep{jiang2024mixtral} and DeepSeek-V3 \citep{liu2024deepseek} operate at high sparsity, activating only a small number of experts per token from a large pool.

\subsection{Connections to Previous Work}

Bayesian treatments of mixtures-of-experts date back to \citet{peng1996bayesian}, who developed MCMC methods enabling uncertainty quantification over expert allocations, while \citet{jiang1999hierarchical} established identifiability conditions and convergence rates for hierarchical MoE. The connection between MoE-like switching structures and state-space modeling was explored by \citet{stroud2003nonlinear} through MCMC algorithms for nonlinear models with state-dependent variances.

Our sparsity mechanism builds on the horseshoe prior \citep{carvalho2010horseshoe} and its global-local shrinkage theory \citep{polson2012half}, with posterior concentration results for nearly-black signals \citep{vanderpas2014horseshoe} and horseshoe-like extensions for subset selection \citep{bhadra2021horseshoelike}. For inference, we build on particle learning for general mixtures \citep{carvalho2010particlea} and interacting particle system theory \citep{johannes2008interacting}.

From the perspective of modern large language models, sparse MoE layers rely on deterministic top-$k$ routing \citep{shazeer2017outrageously, fedus2022switch} with additional objectives for load balancing and diversity. Our contribution is a Bayesian routing alternative that induces sparsity through shrinkage and provides uncertainty quantification and online updating. Finally, approximation theory \citep{zeevi1998error} and recent generalization bounds for sparse MoE \citep{zhao2024generalization} provide statistical context for why sparsity can improve performance when $k\ll K$.

\subsection{Contributions}

This paper introduces \emph{Horseshoe Mixture-of-Experts} (HS-MoE), combining horseshoe priors \citep{carvalho2010horseshoe} with MoE for adaptive sparse expert selection. Our contributions are a hierarchical Bayesian formulation integrating horseshoe priors with MoE, a particle learning algorithm for sequential inference based on sufficient statistics, theoretical connections to approximation and generalization bounds, and an interpretation of HS-MoE as a Bayesian router compatible with modern transformer MoE layers.

The rest of the paper is outlined as follows. Section \ref{sec:hmoe} presents the HS-MoE model formulation and the horseshoe prior. Section \ref{sec:inference} describes the particle learning algorithm for sequential inference. Section \ref{sec:app} details applications to transformer architectures and streaming data. Section \ref{sec:theory} provides theoretical results on approximation and generalization. Section \ref{sec:discussion} concludes with a discussion and comparison of routing approaches.

\section{Horseshoe Mixture of Experts}\label{sec:hmoe}
Let $X \in \mathbb{R}^d$ denote the input and $y$ the response. The MoE model with $K$ experts is:
\begin{equation}\label{eq:moe}
p(Y \mid X, \phi) = \sum_{k=1}^{K} g_k(X; \phi) \, f_k(Y \mid X; \theta_k),
\end{equation}
where $\Theta = \{\phi, \theta_1, \ldots, \theta_K\}$ collects all parameters. The gating function uses the softmax form:
\begin{equation}\label{eq:gating}
g_k(X; \phi) = \frac{\exp(\phi_k^\top X)}{\sum_{l=1}^{K} \exp(\phi_l^\top X)}.
\end{equation}

We introduce latent allocations $z_i \in \{1,\ldots,K\}$ so that the model becomes conditionally independent given $z_{1:n}$:
\begin{align}
y_i \mid X_i, z_i=k, \Theta &\sim f_k(\,\cdot \mid X_i; \theta_k), \\
P(z_i = k \mid X_i, \phi) &= g_k(X_i; \phi),
\end{align}
where $z_i \in \{1, \ldots, K\}$ is the latent expert assignment.

To make the paper self-contained and implementable, we focus on the canonical MoE regression setting with Gaussian linear experts:
\begin{equation}\label{eq:expert-linear}
y_i \mid z_i=k, X_i, \beta_k, \sigma_k^2 \sim \mathcal{N}(X_i^\top \beta_k, \sigma_k^2), \qquad k=1,\ldots,K,
\end{equation}
where $\beta_k \in \mathbb{R}^d$ and $\sigma_k^2>0$. Extensions to generalized linear experts can be handled by the same variance-mean mixture machinery (Section~\ref{sec:inference}, \S\ref{sec:da}).

For each expert we use the Normal-inverse-gamma prior,
\begin{align}
\beta_k \mid \sigma_k^2 &\sim \mathcal{N}(m_{0}, \sigma_k^2 V_{0}), \\
\sigma_k^2 &\sim \text{IG}(a_{0}, b_{0}),
\end{align}
with $V_0 \in \mathbb{R}^{d\times d}$ positive definite. Throughout, $\text{IG}(a,b)$ denotes the inverse-gamma distribution with density $p(x)\propto x^{-(a+1)}\exp(-b/x)$ on $x>0$ (shape $a$ and scale $b$ for $1/x$). This yields closed-form posterior and posterior-predictive updates required by particle learning.

The softmax gate in \eqref{eq:gating} is over-parameterized since adding a common offset to all $\phi_k$ leaves $g_k$ unchanged; a standard identifiability constraint is $\phi_K=0$ (baseline class). In the inference section we instead adopt a stick-breaking logistic gate (Section~\ref{sec:gate}), which is a convenient alternative categorical parameterization that admits conditionally Gaussian updates under P\'olya--Gamma augmentation.

\subsection{Horseshoe Prior for Sparse Expert Selection}

The horseshoe prior \citep{carvalho2010horseshoe, polson2012half} induces sparsity through local-global shrinkage. For each gating coefficient $\phi_{k,j}$ (expert $k$, feature $j$) we set
\begin{align}
\phi_{k,j} \mid \lambda_{k,j}, \tau &\sim \mathcal{N}(0, \tau^2 \lambda_{k,j}^2), \quad k = 1,\ldots,K-1,\ \ j=1,\ldots,d, \\
\lambda_{k,j} &\sim \mathcal{C}^+(0, 1), \\
\tau &\sim \mathcal{C}^+(0, 1),
\end{align}
where $\mathcal{C}^+(0,1)$ is the half-Cauchy distribution. The local shrinkage $\lambda_k$ allows individual coefficients to escape shrinkage when supported by data, while global shrinkage $\tau$ adapts to overall sparsity.

For computational tractability, we use a scale-mixture representation. One convenient form is the inverse-gamma mixture (written here for the local scales):
\begin{equation}
\lambda_{k,j}^2 \mid \nu_{k,j} \sim \text{IG}\!\left(\tfrac{1}{2}, \tfrac{1}{\nu_{k,j}}\right), \quad \nu_{k,j} \sim \text{IG}\!\left(\tfrac{1}{2}, 1\right),
\end{equation}
and similarly for $\tau^2$ (with its own auxiliary variable). Conditional on $(\tau^2,\{\lambda_{k,j}^2\})$, the prior on $\phi_k$ is Gaussian with diagonal covariance $\tau^2 \operatorname{diag}(\lambda_{k,1}^2,\ldots,\lambda_{k,d}^2)$.

\section{Inference via Particle Learning}\label{sec:inference}

We develop particle learning (PL) for sequential Bayesian inference in HS-MoE, following \citet{carvalho2010particlea}. This approach is distinguished by its efficiency in tracking only the essential state vector and by resampling according to predictive probability.

This section proceeds from the general resample--propagate--allocate structure for mixture models (Section~3.2) to the conjugate ingredients required for a practical implementation. We provide closed-form predictive weights and recursive sufficient-statistic updates for Gaussian linear experts (Section~\ref{sec:pred}), and then introduce a categorical gating parameterization (stick-breaking) whose conditional posteriors are Gaussian under P\'olya--Gamma augmentation (Section~\ref{sec:gate}). These components yield a complete particle-learning recursion (Section~\ref{sec:impl}). Finally, we indicate how \citet{polson2012data} extends the approach beyond Gaussian experts via variance--mean mixture augmentation (Section~\ref{sec:da}).

\subsection{Particle Filtering as Interacting Particle System}

A particle filter is a stochastic process run forward in time with marginals equal to the sequence of conditional state posteriors. Let $\mathcal{B}_t$ denote the conditional posterior at time $t$. The key insight from \citet{johannes2008interacting} is that one only needs to simulate forward an interacting particle system whose marginals are precisely $\{\mathcal{B}_1, \mathcal{B}_2, \ldots, \mathcal{B}_t\}$.

Formally, we maintain a population of $N$ particles $\{\mathcal{S}_t^{(i)}, z_{1:t}^{(i)}\}_{i=1}^N$, where $\mathcal{S}_t^{(i)} = (\mathcal{S}_t^{\phi, (i)}, \mathcal{S}_t^{\theta, (i)})$ contains sufficient statistics for gating and experts respectively, and $z_{1:t}^{(i)}$ tracks component allocations. The interacting particle system converges to the true posterior as $N \to \infty$.

\subsection{Algorithm Structure}

Given observations $(X_1, y_1), \ldots, (X_n, y_n)$ arriving sequentially, particle learning proceeds as:

Particles are first \emph{resampled} with weights proportional to the posterior predictive,
\begin{equation}
w_t^{(i)} \propto p(y_t | \mathcal{S}_{t-1}^{(i)}) = \sum_{k=1}^K p(z_t = k | \mathcal{S}_{t-1}^{(i)}) \, p(y_t | z_t = k, \mathcal{S}_{t-1}^{(i)})
\end{equation}
then \emph{propagated} by updating sufficient statistics,
\begin{equation}
\mathcal{S}_t^{(i)} = \left( S^\phi(\mathcal{S}_{t-1}^{\phi, (i)}, X_t, z_t^{(i)}), S^\theta(\mathcal{S}_{t-1}^{\theta, (i)}, y_t, z_t^{(i)}) \right)
\end{equation}
and finally an allocation is drawn from the conditional distribution,
\begin{equation}
z_t^{(i)} \sim p(z_t | y_t, \mathcal{S}_{t-1}^{(i)}) \propto p(z_t | \mathcal{S}_{t-1}^{(i)}) \, p(y_t | z_t, \mathcal{S}_{t-1}^{(i)})
\end{equation}

The resampling step is key: by drawing particles according to predictive probability \emph{before} propagation, we obtain a more efficient set for the next iteration.

\subsection{Closed-form predictive weights for Gaussian linear experts}\label{sec:pred}
For the Gaussian linear expert model \eqref{eq:expert-linear} with the Normal-inverse-gamma prior, each expert's posterior predictive is available in closed form as a Student-$t$. Let $\mathcal{S}_{k,t-1}^{\theta,(i)}$ denote expert $k$'s sufficient statistics in particle $i$ at time $t-1$, corresponding to a Normal-inverse-gamma posterior
\begin{align}
\beta_k \mid \sigma_k^2, \mathcal{S}_{k,t-1}^{\theta,(i)} &\sim \mathcal{N}(m_{k,t-1}^{(i)}, \sigma_k^2 V_{k,t-1}^{(i)}),\\
\sigma_k^2 \mid \mathcal{S}_{k,t-1}^{\theta,(i)} &\sim \text{IG}(a_{k,t-1}^{(i)}, b_{k,t-1}^{(i)}).
\end{align}
Then the one-step predictive is
\begin{equation}\label{eq:student-predictive}
p(y_t \mid z_t=k, X_t, \mathcal{S}_{k,t-1}^{\theta,(i)}) = t_{\nu}\!\left(y_t;\ \mu,\ s^2\right),
\end{equation}
with degrees of freedom $\nu = 2a_{k,t-1}^{(i)}$, location $\mu = X_t^\top m_{k,t-1}^{(i)}$, and scale
\begin{equation}
s^2 = \frac{b_{k,t-1}^{(i)}}{a_{k,t-1}^{(i)}}\left(1 + X_t^\top V_{k,t-1}^{(i)} X_t\right).
\end{equation}
Here $t_{\nu}(y;\mu,s^2)$ denotes the location--scale Student-$t$ density
\[
t_{\nu}(y;\mu,s^2) = \frac{\Gamma((\nu+1)/2)}{\Gamma(\nu/2)\sqrt{\nu\pi s^2}}\left(1+\frac{(y-\mu)^2}{\nu s^2}\right)^{-(\nu+1)/2}.
\]
This is the quantity used inside the mixture predictive $w_t^{(i)}$ in the resampling step.

For completeness, let $P_{k,t} \equiv (V_{k,t})^{-1}$ denote the expert $k$ precision matrix and write $(m_{k,t},P_{k,t},a_{k,t},b_{k,t})$ for the Normal-inverse-gamma posterior parameters within a particle. If $z_t=k$, the conjugate update from $(m_{k,t-1},P_{k,t-1},a_{k,t-1},b_{k,t-1})$ is
\begin{align}
P_{k,t} &= P_{k,t-1} + X_t X_t^\top, \\
m_{k,t} &= P_{k,t}^{-1}\left(P_{k,t-1} m_{k,t-1} + X_t y_t\right), \\
a_{k,t} &= a_{k,t-1} + \tfrac{1}{2}, \\
b_{k,t} &= b_{k,t-1} + \tfrac{1}{2}\left(y_t^2 + m_{k,t-1}^\top P_{k,t-1} m_{k,t-1} - m_{k,t}^\top P_{k,t} m_{k,t}\right).
\end{align}
If $z_t\neq k$, then $(m_{k,t},P_{k,t},a_{k,t},b_{k,t})=(m_{k,t-1},P_{k,t-1},a_{k,t-1},b_{k,t-1})$. These rank-one updates are what we denote abstractly by $S^\theta(\cdot)$ in Algorithm~1.

\subsection{Conditionally conjugate gating updates via stick-breaking P\'olya--Gamma}\label{sec:gate}
The softmax gate \eqref{eq:gating} is a multinomial logistic model. For implementation we adopt a \emph{stick-breaking} logistic gate, which defines a valid $K$-category distribution and yields conditionally Gaussian updates under P\'olya--Gamma augmentation \citep{polson2013bayesian,linderman2015dependent}. This choice is motivated by computational tractability (closed-form Gaussian updates) rather than exact equivalence to the softmax link.

Define $K-1$ binary ``sticks'' with logits $\eta_k(X) = X^\top \phi_k$, $k=1,\ldots,K-1$, and $\sigma(u) = 1/(1+e^{-u})$. The categorical probabilities are
\begin{align}
P(z=k \mid X) &= \sigma(\eta_k(X))\prod_{\ell<k}\bigl(1-\sigma(\eta_\ell(X))\bigr), \quad k=1,\ldots,K-1, \\
P(z=K \mid X) &= \prod_{\ell=1}^{K-1}\bigl(1-\sigma(\eta_\ell(X))\bigr).
\end{align}
This defines a valid $K$-category gate with parameters $\{\phi_1,\ldots,\phi_{K-1}\}$ and is commonly used for Bayesian sequential inference.

Given an allocated class $z_t$, define for each stick $k$ the binary label
\begin{equation}
y_{t,k}^{(z)} \equiv \mathbb{I}(z_t = k), \qquad \text{used only when } z_t \ge k \text{ (otherwise the stick is not visited).}
\end{equation}
Conditional on the event $z_t \ge k$, we have a Bernoulli likelihood
\begin{equation}
y_{t,k}^{(z)} \mid X_t, \phi_k \sim \text{Bernoulli}(\sigma(X_t^\top \phi_k)).
\end{equation}

For each visited stick $k$ at time $t$, introduce $\omega_{t,k} \sim \text{PG}(1, X_t^\top \phi_k)$ and set $\kappa_{t,k}= y_{t,k}^{(z)} - 1/2$ \citep{polson2013bayesian}. Conditional on $\omega_{t,k}$, the contribution of observation $t$ to the (augmented) log-likelihood is quadratic in $\phi_k$:
\begin{equation}
\log p(y_{t,k}^{(z)} \mid X_t,\phi_k,\omega_{t,k}) = \kappa_{t,k} X_t^\top \phi_k - \tfrac{1}{2}\omega_{t,k}(X_t^\top \phi_k)^2 + \text{const}.
\end{equation}
Thus, conditional on horseshoe scales $(\tau^2,\lambda_{k,1:d}^2)$, the posterior for $\phi_k$ is Gaussian with precision and mean updated by rank-one increments:
\begin{align}
\Lambda_{k,t} &= \Lambda_{k,t-1} + \omega_{t,k} X_t X_t^\top,\\
h_{k,t} &= h_{k,t-1} + \kappa_{t,k} X_t,
\end{align}
where the prior contributes $\Lambda_{k,0} = (\tau^2 \operatorname{diag}(\lambda_{k,1:d}^2))^{-1}$ and $h_{k,0}=0$. The implied Gaussian posterior is $\phi_k \mid \cdot \sim \mathcal{N}(m_{k,t}, \Lambda_{k,t}^{-1})$ with $m_{k,t} = \Lambda_{k,t}^{-1} h_{k,t}$.

\subsection{Implementation-ready particle learning loop}\label{sec:impl}
Each particle stores expert sufficient statistics for $k=1,\ldots,K$ (e.g., $(m_{k,t},V_{k,t},a_{k,t},b_{k,t})$ for the Normal-inverse-gamma posterior), gating sufficient statistics for sticks $k=1,\ldots,K-1$ (e.g., $(\Lambda_{k,t},h_{k,t})$), the current horseshoe scales $(\tau^2,\lambda_{k,1:d}^2)$, and a current draw (or posterior mean) of each stick coefficient vector $\phi_k$, together with the current allocation history. At time $t$, we compute $w_t$ using \eqref{eq:student-predictive}, resample particles, sample $z_t$ from the categorical posterior proportional to $P(z_t=k\mid X_t,\phi)\,p(y_t\mid z_t=k,\mathcal{S}_{k,t-1}^\theta)$, update the chosen expert's statistics, and update visited sticks via P\'olya--Gamma augmentation, after which $\phi_k$ is refreshed by sampling $\phi_k \sim \mathcal{N}(m_{k,t},\Lambda_{k,t}^{-1})$; alternatively, one may use the posterior mean $\phi_k\leftarrow m_{k,t}$. One may also include rejuvenation steps for the horseshoe scales.

\subsection{Marginal Likelihood and Model Selection}

Particle learning provides straightforward marginal likelihood estimation:
\begin{equation}
\hat{p}(y_{1:n}) = \prod_{t=1}^n \left( \frac{1}{N} \sum_{i=1}^N p(y_t | \mathcal{S}_{t-1}^{(i)}) \right),
\end{equation}
which enables Bayesian model comparison across different numbers of experts $K$. This is a significant advantage over MCMC, where marginal likelihood estimation requires additional computational effort.

\begin{algorithm}[H]
\caption{Particle Learning for HS-MoE}
\KwIn{Observations $(X_1, y_1), \ldots, (X_n, y_n)$, number of particles $N$, experts $K$}
\KwOut{Particle approximation to posterior, marginal likelihood estimate}
Initialize particles $\{\mathcal{S}_0^{(i)}\}_{i=1}^N$ from prior\;
$\hat{p}(y_{1:n}) \leftarrow 1$\;
\For{$t = 1, \ldots, n$}{
  \For{$i = 1, \ldots, N$}{
    Compute predictive weight $w_t^{(i)} = p(y_t | \mathcal{S}_{t-1}^{(i)})$\;
  }
  $\hat{p}(y_{1:n}) \leftarrow \hat{p}(y_{1:n}) \times \frac{1}{N}\sum_{i=1}^N w_t^{(i)}$\;
  Resample indices $\{a_t^{(i)}\}$ with probabilities $\propto w_t^{(i)}$\;
  \For{$i = 1, \ldots, N$}{
    Draw $z_t^{(i)} \sim p(z_t | y_t, \mathcal{S}_{t-1}^{(a_t^{(i)})})$\;
    Update expert sufficient statistics $\mathcal{S}_t^{\theta,(i)} = S^\theta(\mathcal{S}_{t-1}^{\theta,(a_t^{(i)})}, X_t, y_t, z_t^{(i)})$\;
    \For{$k = 1,\ldots,K-1$}{
      \If{$z_t^{(i)} \ge k$}{
        Set $y_{t,k}^{(z)} \leftarrow \mathbb{I}(z_t^{(i)} = k)$ and $\kappa_{t,k} \leftarrow y_{t,k}^{(z)} - 1/2$\;
        Draw $\omega_{t,k}^{(i)} \sim \text{PG}(1, X_t^\top \phi_k^{(i)})$\;
        Update gating sufficient statistics $\mathcal{S}_t^{\phi,(i)}$ via $\Lambda_{k,t} \leftarrow \Lambda_{k,t-1} + \omega_{t,k}^{(i)} X_t X_t^\top$, $h_{k,t} \leftarrow h_{k,t-1} + \kappa_{t,k} X_t$\;
        Refresh $\phi_k^{(i)} \sim \mathcal{N}(m_{k,t}^{(i)}, \Lambda_{k,t}^{(i)\,-1})$ (or set $\phi_k^{(i)} \leftarrow m_{k,t}^{(i)}$)\;
      }
    }
  }
}
\Return{$\{\mathcal{S}_n^{(i)}, z_{1:n}^{(i)}\}_{i=1}^N$, $\hat{p}(y_{1:n})$}
\end{algorithm}

\subsection{Computational Complexity}

For Gaussian linear experts with $d$ features, the per-observation cost is dominated by evaluating $K$ expert predictives for each particle and updating one expert and up to $K-1$ sticks. With dense linear algebra, this is $O(N K d^2)$ per observation (due to rank-one updates and linear solves). In implementations, maintaining Cholesky factorizations for the relevant precision matrices improves numerical stability and reduces constants. This compares favorably to batch MCMC, which requires $O(nK d^2)$ per iteration and many iterations for convergence. For streaming data, particle learning processes each observation once, achieving $O(n N K d^2)$ total complexity versus $O(T n K d^2)$ for MCMC with $T$ iterations.

\subsection{Data Augmentation, Conditionally Conjugate Sufficient Statistics}\label{sec:da}

Particle learning requires that (conditional) predictive probabilities and parameter updates be available in closed form. For Gaussian experts \eqref{eq:expert-linear} this is already true (Section~\ref{sec:pred}). For non-Gaussian expert likelihoods or robust losses, \citet{polson2012data} provide a general \emph{variance--mean mixture} representation that makes a broad class of models conditionally Gaussian given latent variables $\omega$.

Concretely, for many likelihoods (or pseudo-likelihoods) with linear predictor $\eta_i = X_i^\top \theta$, one can introduce $\omega_i$ such that
\begin{equation}\label{eq:vm-mixture}
p(y_i \mid \eta_i) \propto \int \exp\!\Big(\kappa(y_i)\,\eta_i - \tfrac{1}{2}\omega_i \eta_i^2\Big)\, p(\omega_i \mid y_i)\, d\omega_i,
\end{equation}
so that conditional on $\omega_i$, the dependence on $\theta$ is Gaussian (quadratic in $\eta_i$). With a Gaussian prior on $\theta$, the conditional posterior is Gaussian and can be summarized by sufficient statistics (the same $\Lambda,h$ structure as in Section~\ref{sec:gate}). This is the mechanism by which particle learning extends beyond purely Gaussian experts.

For example, the Laplace likelihood $p(y\mid \eta)\propto \exp(-|y-\eta|/b)$ admits a scale-mixture-of-Gaussians representation with an exponential mixing distribution on the variance; conditional on the mixing variable, $(y\mid \eta,\omega)$ is Gaussian \citep{polson2012data}, yielding conjugate updates for robust expert regressions within each particle.

\subsection{Alternative: MCMC with Data Augmentation}

For batch inference, standard MCMC provides an alternative using data augmentation \citep{polson2012data}. The inverse-gamma representation enables Gibbs sampling with conjugate conditionals. However, MCMC requires processing all data each iteration, making particle learning preferable for streaming or large-scale settings.

\subsection{P\'olya--Gamma identity (logistic gating)}

For the Bernoulli logit likelihood (used by each stick in Section~\ref{sec:gate}), the P\'olya--Gamma identity \citep{polson2013bayesian} gives an exact augmentation:
\begin{equation}
\frac{(e^\psi)^a}{(1 + e^\psi)^b} = 2^{-b} e^{\kappa \psi} \int_0^\infty e^{-\omega \psi^2/2} p(\omega) d\omega,
\end{equation}
where $\omega \sim \text{PG}(b, 0)$. Conditional on $\omega$, the logit likelihood becomes quadratic in $\psi$, hence Gaussian in the regression coefficients. Stick-breaking extends this construction from binary logits to categorical gating while preserving conditional Gaussian updates \citep{linderman2015dependent}.

\subsection{Integration with Transformer Architectures}

In modern transformer architectures, MoE layers replace the feed-forward network (FFN) in selected blocks. Given an input token representation $X$, the HS-MoE layer computes:
\begin{equation}
\text{HS-MoE}(X) = \sum_{k=1}^{K} g_k(X; \phi) \, \text{FFN}_k(X),
\end{equation}
where horseshoe priors on the gating parameters $\phi$ induce adaptive sparsity. Each FFN expert has the standard form:
\begin{equation}
\text{FFN}_k(X) = W_k^{(2)} \, \sigma(W_k^{(1)} X + b_k^{(1)}) + b_k^{(2)},
\end{equation}
where $\sigma(\cdot)$ is a nonlinear activation (GeLU or SiLU).

\begin{remark}[Comparison to Top-$k$ Routing]
Standard sparse MoE uses hard top-$k$ selection, activating exactly $k$ experts per token regardless of input. In contrast, HS-MoE provides soft sparsity where the effective number of experts is data-driven. When most $\lambda_k \approx 0$, the posterior concentrates on configurations with few active experts, but the model can activate more experts when the data warrants it.
\end{remark}

\section{Application}\label{sec:app}

Training and deployment of MoE layers in large language models confront expert collapse (a small subset of experts dominates), routing sensitivity under distribution shift, and strict compute constraints that require top-$k$ activation per token. HS-MoE addresses these issues by placing structured global-local shrinkage on router parameters, encouraging most expert logits to be effectively inactive while permitting a small subset to escape shrinkage when supported by data, and by representing uncertainty over routing decisions.

Despite its probabilistic formulation, HS-MoE is compatible with compute-constrained inference. In deployment, one uses the Bayesian router to produce expert scores and then applies standard top-$k$ selection: experts may be ranked by posterior mean logits $\mathbb{E}[\eta_k(X)\mid \mathcal{D}]$ (or a posterior draw), with optional uncertainty-aware ranking such as 
\[
\mathbb{E}[\eta_k(X)]-\alpha\sqrt{\mathrm{Var}(\eta_k(X))},
\] 
to avoid unstable routing early in training or under distribution shift. The resulting expert selection and compute budget are identical to standard sparse MoE inference; only the scoring function used to rank experts is modified.

Token-level particle learning is generally too expensive for large-scale pretraining, so HS-MoE should be viewed as a Bayesian router that can be trained with scalable approximations. One option is variational Bayes for the router, maintaining an approximate posterior over $\phi$ and horseshoe scales and using minibatches of tokens with P\'olya--Gamma (or a local quadratic approximation) to obtain stable, approximately Gaussian updates. A second option is MAP-style training with horseshoe-like regularization on router weights, optimized by SGD/Adam jointly with expert parameters, which preserves the sparse routing inductive bias while matching standard training pipelines. A practical hybrid is to train experts with standard optimization and periodically re-fit the router (and its uncertainty) on recent activations using the sequential updates in Sections~\ref{sec:gate}--\ref{sec:impl}.

Particle learning is most viable in streaming and continual settings such as domain adaptation, personalization, and interaction logs, where data arrive sequentially and the router must adapt without reprocessing the full history. In these regimes, posterior predictive resampling provides a principled mechanism for online routing adaptation with uncertainty quantification. Evaluation may report perplexity or downstream accuracy, expert utilization and load balance, routing entropy and calibration, robustness under distribution shift, and training stability (including router/expert collapse frequency).

\subsection{Example: Gaussian Mixture Regression}

To illustrate the mechanics of HS-MoE in a setting where all expert updates are closed form, we include a synthetic regression example with $K=10$ Gaussian linear experts \eqref{eq:expert-linear} and $n=500$ observations, where only $s=3$ experts are active. Covariates are generated as $X_i \sim \mathcal{N}(0,I_d)$ with $d=5$, and allocations $z_i$ are drawn from a sparse data-generating gate so that $P(z_i\in\{1,2,3\}\mid X_i)$ dominates. Conditional on $z_i=k$, responses are sampled as $y_i = X_i^\top \beta_k + \varepsilon_i$ with $\varepsilon_i\sim\mathcal{N}(0,\sigma_k^2)$. The reproducible setup and outputs shown below are generated by running \texttt{python scripts/generate\_synth\_example.py}, which writes the LaTeX tables to \texttt{generated/} and the figure to \texttt{fig/}.

In applications of particle learning (Section~\ref{sec:impl}), posterior mean allocation frequencies are computed by averaging particle-averaged allocation probabilities across time. For the synthetic generator we report the empirical allocation frequencies from the data-generating allocations, which serve as a reference target for what a well-calibrated sparse router should recover.

\begin{table}[h]
\centering
\caption{Synthetic HS-MoE regression example (reproducible setup).}
\label{tab:synth-setup}
\begin{tabular}{lc}
\hline
\textbf{Quantity} & \textbf{Value} \\
\hline
Number of experts $K$ & 10 \\
Active experts $s$ & 3 (experts 1--3) \\
Sample size $n$ & 500 \\
Feature dimension $d$ & 5 \\
Particles $N$ & 1000 \\
Experts & Gaussian linear, Normal--IG prior \\
Gate & softmax, sparse by construction ($b_{\mathrm{inactive}}=-3.0$, $T=0.70$) \\
\hline
\end{tabular}
\end{table}

\begin{figure}[h]
\centering
\includegraphics[width=0.82\linewidth]{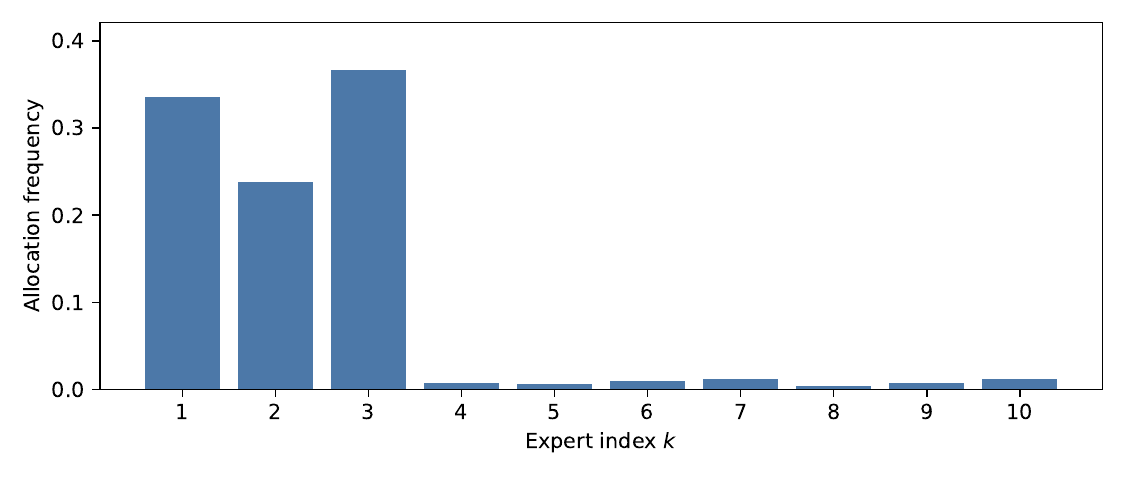}
\caption{Allocation frequencies across experts in the synthetic example (generated by \texttt{scripts/generate\_synth\_example.py}).}
\label{fig:synth-alloc}
\end{figure}

\section{Theoretical Properties}\label{sec:theory}

Mixture-of-experts models achieve universal approximation with favorable rates. For functions in the Sobolev class $W_p^r$ on $[0,1]^d$, MoE with $K$ experts achieves approximation error \citep{zeevi1998error}:
\begin{equation}
\inf_{f_{\text{MoE}} \in \mathcal{M}_K} \|f - f_{\text{MoE}}\|_{L_p} = O\left( K^{-r/d} \right),
\end{equation}
where $r$ is the smoothness parameter. This rate matches that of free-knot splines and neural networks.

For sparse MoE activating $k$ of $K$ experts, recent work \citep{zhao2024generalization} establishes generalization error bounds depending on the Rademacher complexity $\mathcal{R}_n(\mathcal{H})$ of the expert class and the Natarajan dimension $d_N$ of the router:
\begin{equation}
\mathcal{E}_{\text{gen}} = O\left( \mathcal{R}_n(\mathcal{H}) + \sqrt{\frac{k \cdot d_N (1 + \log(K/k))}{n}} \right).
\end{equation}
This bound explicitly shows how sparsity ($k \ll K$) improves generalization. The horseshoe prior provides a soft mechanism for achieving sparsity, where the effective $k$ is determined by the posterior rather than a hard constraint.

\section{Discussion}\label{sec:discussion}

We have introduced Horseshoe Mixture-of-Experts (HS-MoE), a Bayesian framework combining three key elements: the horseshoe prior's adaptive shrinkage for automatic expert selection \citep{carvalho2010horseshoe, vanderpas2014horseshoe}, particle learning for efficient sequential inference \citep{carvalho2010particlea}, and interacting particle system theory for convergence guarantees \citep{johannes2008interacting}. The particle learning framework offers several advantages over batch MCMC, including sequential processing of observations, reduced storage requirements through sufficient statistic tracking, natural handling of streaming data, and straightforward marginal likelihood computation for model selection. While stochastic gradient descent is standard for training MoE in deep learning, our approach offers a probabilistic alternative that quantifies uncertainty and enables online model selection, albeit at higher computational cost per sample.

Compared to existing routing mechanisms (Table~\ref{tab:comparison}), HS-MoE provides adaptive soft sparsity rather than the hard top-$k$ selection or the dense computation of Soft MoE. The number of active experts is data-driven rather than fixed, and the Bayesian framework enables principled uncertainty quantification and model selection via marginal likelihoods rather than heuristics.

\begin{table}[h]
\centering
\caption{Comparison of expert routing methods}
\label{tab:comparison}
\begin{tabular}{lccc}
\hline
\textbf{Property} & \textbf{Top-$k$} & \textbf{Soft MoE} & \textbf{HS-MoE} \\
\hline
Sparsity type & Hard & None & Soft (adaptive) \\
Experts per input & Fixed $k$ & All $K$ & Data-driven \\
Uncertainty quantification & No & No & Yes \\
Sequential inference & No & No & Yes \\
Model selection & Heuristic & Heuristic & Marginal likelihood \\
\hline
\end{tabular}
\end{table}

Several directions merit future investigation: variational approximations \citep{ghosh2017structured} for scaling to transformer-sized models, group-level horseshoe priors for structured layer-wise expert selection, nonparametric priors for online adjustment of the number of experts $K$, and connections to Kolmogorov-Arnold Networks \citep{polson2025kolmogorov} for efficient function approximation. As mixture-of-experts architectures continue to scale in modern LLMs, Bayesian approaches will play an increasingly important role in understanding and improving these systems.

\bibliography{HS-MOE}

\begin{thebibliography}{23}
\providecommand{\natexlab}[1]{#1}
\providecommand{\url}[1]{\texttt{#1}}
\expandafter\ifx\csname urlstyle\endcsname\relax
  \providecommand{\doi}[1]{doi: #1}\else
  \providecommand{\doi}{doi: \begingroup \urlstyle{rm}\Url}\fi

\bibitem[Bhadra et~al.(2021)Bhadra, Datta, Polson, and Willard]{bhadra2021horseshoelike}
Anindya Bhadra, Jyotishka Datta, Nicholas~G. Polson, and Brandon~T. Willard.
\newblock The {{Horseshoe-Like Regularization}} for {{Feature Subset Selection}}.
\newblock \emph{Sankhya B}, 83\penalty0 (1):\penalty0 185--214, May 2021.

\bibitem[Carvalho et~al.(2010{\natexlab{a}})Carvalho, Lopes, Polson, and Taddy]{carvalho2010particlea}
Carlos~M Carvalho, Hedibert~F Lopes, Nicholas~G Polson, and Matt~A Taddy.
\newblock Particle learning for general mixtures.
\newblock \emph{Bayesian Analysis}, 5\penalty0 (4):\penalty0 709--740, 2010{\natexlab{a}}.

\bibitem[Carvalho et~al.(2010{\natexlab{b}})Carvalho, Polson, and Scott]{carvalho2010horseshoe}
Carlos~M. Carvalho, Nicholas~G. Polson, and James~G. Scott.
\newblock The horseshoe estimator for sparse signals.
\newblock \emph{Biometrika}, page asq017, 2010{\natexlab{b}}.

\bibitem[Fedus et~al.(2022)Fedus, Zoph, and Shazeer]{fedus2022switch}
William Fedus, Barret Zoph, and Noam Shazeer.
\newblock Switch {{Transformers}}: {{Scaling}} to {{Trillion Parameter Models}} with {{Simple}} and {{Efficient Sparsity}}.
\newblock \emph{Journal of Machine Learning Research}, 23\penalty0 (120):\penalty0 1--39, 2022.

\bibitem[Ghosh et~al.(2017)Ghosh, Yao, and {Doshi-Velez}]{ghosh2017structured}
Soumya Ghosh, Jialin Yao, and Finale {Doshi-Velez}.
\newblock Model selection in {{Bayesian}} neural networks via horseshoe priors.
\newblock In \emph{Proceedings of the 34th International Conference on Machine Learning}, pages 1229--1238, 2017.

\bibitem[Jacobs et~al.(1991)Jacobs, Jordan, Nowlan, and Hinton]{jacobs1991adaptive}
Robert~A. Jacobs, Michael~I. Jordan, Steven~J. Nowlan, and Geoffrey~E. Hinton.
\newblock Adaptive {{Mixtures}} of {{Local Experts}}.
\newblock \emph{Neural Computation}, 3\penalty0 (1):\penalty0 79--87, 1991.

\bibitem[Jiang et~al.(2024)Jiang, Sablayrolles, Roux, Mensch, et~al.]{jiang2024mixtral}
Albert~Q Jiang, Alexandre Sablayrolles, Antoine Roux, Arthur Mensch, et~al.
\newblock Mixtral of experts.
\newblock \emph{arXiv preprint arXiv:2401.04088}, 2024.

\bibitem[Jiang and Tanner(1999)]{jiang1999hierarchical}
Wenxin Jiang and Martin~A Tanner.
\newblock Hierarchical mixtures-of-experts for exponential family regression models: Approximation and maximum likelihood estimation.
\newblock \emph{The Annals of Statistics}, 27\penalty0 (3):\penalty0 987--1011, 1999.

\bibitem[Johannes et~al.(2008)Johannes, Polson, and Stroud]{johannes2008interacting}
Michael Johannes, Nicholas Polson, and Jonathan Stroud.
\newblock Interacting {{Particle Systems}} for {{Sequential Parameter Learning}}, 2008.

\bibitem[Jordan and Jacobs(1994)]{jordan1994hierarchical}
Michael~I Jordan and Robert~A Jacobs.
\newblock Hierarchical mixtures of experts and the {{EM}} algorithm.
\newblock \emph{Neural computation}, 6\penalty0 (2):\penalty0 181--214, 1994.

\bibitem[Linderman et~al.(2015)Linderman, Johnson, and Adams]{linderman2015dependent}
Scott~W. Linderman, Matthew~J. Johnson, and Ryan~P. Adams.
\newblock Dependent multinomial models made easy: Stick-breaking with the {P{\'o}lya--Gamma} augmentation.
\newblock In \emph{Advances in Neural Information Processing Systems}, 2015.

\bibitem[Liu et~al.(2024)Liu, Feng, Xue, Wang, et~al.]{liu2024deepseek}
Aixin Liu, Bei Feng, Bin Xue, Bingxuan Wang, et~al.
\newblock {{DeepSeek-V3}} technical report.
\newblock \emph{arXiv preprint arXiv:2412.19437}, 2024.

\bibitem[Peng et~al.(1996)Peng, Jacobs, and Tanner]{peng1996bayesian}
Fengchun Peng, Robert~A. Jacobs, and Martin~A. Tanner.
\newblock Bayesian inference in mixtures-of-experts and hierarchical mixtures-of-experts models with an application to speech recognition.
\newblock \emph{Journal of the American Statistical Association}, 91\penalty0 (435):\penalty0 953--960, 1996.

\bibitem[Polson and Scott(2012{\natexlab{a}})]{polson2012data}
Nicholas~G. Polson and James~G. Scott.
\newblock Data augmentation for non-{{Gaussian}} regression models using variance-mean mixtures, September 2012{\natexlab{a}}.

\bibitem[Polson and Scott(2012{\natexlab{b}})]{polson2012half}
Nicholas~G Polson and James~G Scott.
\newblock Half-{{Cauchy}} priors for scale parameters and global-local shrinkage priors.
\newblock \emph{Bayesian Analysis}, 7\penalty0 (4):\penalty0 887--902, 2012{\natexlab{b}}.

\bibitem[Polson et~al.(2013)Polson, Scott, and Windle]{polson2013bayesian}
Nicholas~G Polson, James~G Scott, and Jesse Windle.
\newblock Bayesian inference for logistic models using {{P\'olya-Gamma}} latent variables.
\newblock \emph{Journal of the American Statistical Association}, 108\penalty0 (504):\penalty0 1339--1349, 2013.

\bibitem[Polson and Sokolov(2025)]{polson2025kolmogorov}
Sarah Polson and Vadim Sokolov.
\newblock Kolmogorov {{GAM Networks Are All You Need}}!
\newblock \emph{Entropy}, 27\penalty0 (6):\penalty0 593, June 2025.

\bibitem[Shazeer et~al.(2017)Shazeer, Mirhoseini, Maziarz, Davis, Le, Hinton, and Dean]{shazeer2017outrageously}
Noam Shazeer, Azalia Mirhoseini, Krzysztof Maziarz, Andy Davis, Quoc Le, Geoffrey Hinton, and Jeff Dean.
\newblock Outrageously {{Large Neural Networks}}: {{The Sparsely-Gated Mixture-of-Experts Layer}}.
\newblock In \emph{International {{Conference}} on {{Learning Representations}}}, 2017.

\bibitem[Stroud et~al.(2003)Stroud, M{\"u}ller, and Polson]{stroud2003nonlinear}
Jonathan~R. Stroud, Peter M{\"u}ller, and Nicholas~G. Polson.
\newblock Nonlinear {{State-Space Models}} with {{State-Dependent Variances}}.
\newblock \emph{Journal of the American Statistical Association}, 98\penalty0 (462):\penalty0 377--386, 2003.

\bibitem[Titterington et~al.(1985)Titterington, Smith, and Makov]{titterington1985statistical}
D.~M. (David~Michael) Titterington, Adrian F.~M. Smith, and U.~E. Makov.
\newblock \emph{Statistical Analysis of Finite Mixture Distributions}.
\newblock Wiley Series in Probability and Mathematical Statistics. Wiley, 1985.

\bibitem[{van der Pas} et~al.(2014){van der Pas}, Kleijn, and {van der Vaart}]{vanderpas2014horseshoe}
S.~L. {van der Pas}, B.~J.~K. Kleijn, and A.~W. {van der Vaart}.
\newblock The horseshoe estimator: {{Posterior}} concentration around nearly black vectors.
\newblock \emph{Electronic Journal of Statistics}, 8\penalty0 (2):\penalty0 2585--2618, 2014.

\bibitem[Zeevi et~al.(1998)Zeevi, Meir, and Maiorov]{zeevi1998error}
Assaf~J Zeevi, Ron Meir, and Vitaly Maiorov.
\newblock Error bounds for functional approximation and estimation using mixtures of experts.
\newblock \emph{IEEE Transactions on Information Theory}, 44\penalty0 (3):\penalty0 1010--1025, 1998.

\bibitem[Zhao et~al.(2024)Zhao, Wang, and Wang]{zhao2024generalization}
Yankai Zhao, Huayan Wang, and Wenjie Wang.
\newblock Generalization error analysis for sparse mixture-of-experts: {{A}} preliminary study.
\newblock \emph{arXiv preprint arXiv:2403.17404}, 2024.

\end{thebibliography}

\end{document}